\documentclass[]{svmult}

\usepackage{mathptmx}       
\usepackage{helvet}         
\usepackage{courier}        
\usepackage{type1cm}        
\usepackage{makeidx}         
\usepackage{graphicx}        
\usepackage{multicol}        
\usepackage[bottom]{footmisc}

\usepackage{titlesec}

\titleformat{\section}[hang]
  {\normalfont\Large\bfseries}{\thesection}{1em}{}
\titlespacing*{\section}
  {0pt}{3.5ex plus 1ex minus .2ex}{2.3ex plus .2ex}

\usepackage{caption}
\makeindex             

\begin{document}

\title*{Multi-Intent Recognition in Dialogue Understanding:  A Comparison Between Smaller Open-Source LLMs}
\titlerunning{Multi-Intent Recognition in Dialogue Understanding}
\author{Adnan Ahmad and Philine Kowol and Stefan Hillmann and Sebastian Möller }
\institute{Adnan Ahmad \at Technische Universität Berlin, 10623 Berlin, Germany, \email{adnan.ahmad@tu-berlin.de}
\and Philine Kowol \at Technische Universität Berlin, 10623 Berlin, Germany, \email{p.goerzig@tu-berlin.de}
\and Stefan Hillmann \at Technische Universität Berlin, 10623 Berlin, Germany, \email{stefan.hillmann@tu-berlin.de}
\and Sebastian Möller \at Technische Universität Berlin, 10623 Berlin, Germany, \email{sebastian.moeller@tu-berlin.de}}

%
%
\maketitle

\abstract{In this paper, we provide an extensive analysis of multi-label intent classification using Large Language Models (LLMs) that are open-source, publicly available, and can be run in consumer hardware. We use the MultiWOZ 2.1 dataset, a benchmark in the dialogue system domain, to investigate the efficacy of three popular open-source pre-trained LLMs, namely \textbf{LLama2-7B-hf}, \textbf{Mistral-7B-v0.1}, and \textbf{Yi-6B}. We perform the classification task in a few-shot setup, giving 20 examples in the prompt with some instructions. Our approach focuses on the differences in performance of these models across several performance metrics by methodically assessing these models on multi-label intent classification tasks. Additionally, we compare the performance of the instruction-based fine-tuning approach with supervised learning using the smaller transformer model BertForSequenceClassification as a baseline. To evaluate the performance of the models, we use evaluation metrics like accuracy, precision, and recall as well as micro, macro, and weighted F1 score. We also report the inference time, VRAM requirements, etc. The Mistral-7B-v0.1 outperforms two other generative models on 11 intent classes out of 14 in terms of \textbf{F-Score}, with a weighted average of \textbf{0.50}. It also has relatively lower Humming Loss and higher Jaccard Similarity, making it the winning model in the few-shot setting. We find BERT based supervised classifier having superior performance compared to the best performing few-shot generative LLM. The study provides a framework for small open-source LLMs in detecting complex multi-intent dialogues, enhancing the Natural Language Understanding aspect of task-oriented chatbots.}

\section{Introduction}
\label{sec:intro}
Recent advancements in the domain of Large Language Models and Conversational AI have been showing tremendous potential in few-shot learning~\cite{fewshot}. The sheer amount of data and model parameters leads to the model's emerging capability of performing unknown tasks with little to no amount of training data. This has been rigorously reported on many benchmark natural language processing tasks, for example, question answering~\cite{llm-qa}, document summarization~\cite{llm-sam}, information extraction~\cite{llm-ie}, and language generation~\cite{llm-nlg} etc. However critical problems remain about the size and availability of the models for them to use in a consumer hardware setup without sharing the confidential data. Most successful LLMs are proprietary models and need to be accessed via commercial APIs\footnote{https://openai.com/gpt-4}\textsuperscript{,}\footnote{https://www.anthropic.com/index/introducing-claude}\textsuperscript{,}\footnote{https://blog.google/technology/ai/google-palm-2-ai-large-language-model/}. The massive size of these models is not convenient to host locally. For example, GPT-3 has 175B parameters~\cite{llm-sam}, PALM-2 has 540B parameters~\cite{llm-qa}, Claude-2 has 137 billion parameter~\cite{claude1,claude2}, etc. Therefore they are barely being used to deal with confidential data, or data that requires higher privacy, as they often collect user’s data to farther train the model\footnote{https://openai.com/enterprise-privacy} and not doing so requires enterprise licensing. Thankfully, a growing number of small to medium size open-source large language models (\textit{somoLLMs}) have been published recently. Such somoLLMs can be downloaded to local machines and be used in consumer hardware efficiently. 

Some prominent somoLLMs are LLama 2~\cite{llama-2}, Mistral~\cite{mistral}, and Yi~\cite{yi}. However, smaller models often come with the cost of reduced ability. So they need to be evaluated on different tasks before using them. Their performance depend on the task, model’s choice, model size, user prompt, pre-trained vs fine-tuned, and so on. So instead of relying on evaluation of some generic tasks, it is always important to evaluate the models on the task at hand. While doing so, it is often best practice to compare the performance of multiple models on the task and choose the best one. One such task is an intent classification for multi-turn dialogues~\cite{intent1}. 

In this work, we use MultiWoz~2.1~\cite{multiwoz1} dataset, a widely used human annotated multi-turn dialogue dataset, three small-scale pre-trained somoLLMs with about 7B parameters to perform few shot dialogue intent classification, and a BERT-based model~\cite{devlin-etal-2019-bert}  for supervised learning using BERT-embeddings as baseline. The somoLLMs that we choose are Llama2-7B-hf, Mistral-7B-v0.1, and Yi-6B, the three top-performing models in the Huggingface leaderboard~\cite{hf-lead} for 7B parameter range. The reason we choose 7B parameter range models as they and their quantized versions can easily fit in today's consumer GPUs. We do not fine-tune these models farther as fine-tuning even such relatively small models requires substantial amounts of hardware resource, therefore can’t be used in of-the shelf manner. For supervised learning baseline, we have selected BertForSequenceClassification\footnote{https://huggingface.co/transformers/v3.5.1/model\_doc/bert.html\#bertforsequenceclassification} as it is optimized for sequential textual data like utterances in a dialog.

As the MultiWoz~2.1 dataset contains multi-label intents, therefore its a multi-label classification task, we present appropriate evaluation metrics in the result section. We only present the results of the user utterance classification and ignore the system utterance results as they are often generated by templates in most dialogue systems. We used validation data to choose our prompt and used the same prompt across different models on the same test data for few-shot learning. More details will be given in the corresponding data and experiment section. We also discuss the model size and inference time. Then we compare the LLM with the best performance in few-shot setup with the BERT based supervised approach. This paper will help the speech and dialogue community with a standard of how to set up such individual tasks and evaluation, and also inform the researchers and developers about the comparative model performance on multi-label intent classification tasks on multi-turn dialogues.

\section{Related Works}
\label{sec:rel_work}
Natural language processing (NLP) has seen a revolution with the introduction of large language models (LLMs). The emergence of auto-regressive Large Language Models such as GPT-3, PALM-2, Claude-2, FLAN-t5 are notable development in this field~\cite{fewshot,llm-qa,claude2}. They have shown exceptional few-shot performance on a range of NLP tasks, such as translation, question-answering, and text-summarising tasks. This capability of LLMs, to perform tasks with little to no training data, opens up new possibilities for conversational systems to achieve effective intent recognition. Recent studies have proposed various techniques like data augmentation for intent classification using off-the-shelf LLMs such as GPT-3, highlighting the models' ability to generate training data without task-specific fine-tuning~\cite{data-aug}.

In recent times, there has been significant progress in the development of smaller, open-source Large Language Models (somoLLMs) that are more accessible and manageable for a wider range of applications~\cite{llama-2,mistral}. These models, often scaled-down versions of their larger counterparts, provide a balance between performance and resource efficiency. They are particularly useful for tasks in environments with limited computational resources or budget constraints. Open-source availability means that researchers and developers can modify and fine-tune these models for specific applications without the prohibitive costs associated with larger models. These smaller somoLLMs, while not matching the full capabilities of their larger counterparts, still offer robust natural language understanding (NLU) and generation (NLG), making them ideal for NLU tasks, especially when integrated with internal, i.e. private or confidential, data and chatbots.

The MultiWOZ~2.1 dataset~\cite{multiwoz1} is a benchmark dataset for dialogue systems. It spans seven distinct domains and contains over 10,000 dialogues, providing a diverse testing ground for intent classification. Although there are many transformer based approaches has been proposed for dialogue state tracking and natural language understanding on this dataset~\cite{intent1,intent2,intent2}, to our best knowledge, there has been no reports on few shot intent classification performance using relatively small somoLLMs. Therefore, we see a need to evaluate his subject.

\section{Dataset}
\label{sec:dataset}

For our study, we use the MultiWOZ~2.1 dataset\footnote{https://github.com/budzianowski/multiwoz/blob/master/data/MultiWOZ\_2.1.zip}~\cite{multiwoz1}, an enhanced version of the original MultiWOZ~2.0~\cite{multiwoz2} dataset, which is widely recognized in conversational AI research. This version addresses several annotation errors and state inconsistencies found in its predecessor, MultiWOZ 2.1, thereby providing a more reliable and coherent dataset. MultiWOZ 2.1 is publicly available and adheres to ethical guidelines for data collection and usage. It is sourced from crowd-sourced dialogues, ensuring a diverse representation of linguistic patterns and user behaviours.

MultiWOZ~2.1 comprises over 10,000 multi-turn dialogues, encompassing seven distinct domains: Attraction, Hotel, Restaurant, Taxi, Train, Hospital, and Police. Each dialogue is meticulously annotated with dialogue acts and spans across various topics, offering a comprehensive landscape for intent classification. The multifaceted nature of the conversations in MultiWOZ~2.1 makes it particularly suitable for multi-label intent classification. The dialogues often exhibit overlapping intents within a single user utterance, reflecting realistic and complex user queries. This characteristic challenges the model to discern and classify multiple intents concurrently, a critical requirement for advanced conversational AI systems~\cite{intent-2}.

\begin{table}[]
\centering
\caption{Dialogue sessions and user turns in Train, Test, and Validation sets}
\label{tab:data_distribution}
\begin{tabular}{lccc}
\hline
\textbf{Set} & \textbf{Dialogue Sessions} & \textbf{User Turns}  & \textbf{Intent Classes}\\ 
\hline
All  & 10,438 & 71,524 & 14\\
Train & 8,438 & 56778 & 14 \\
Validation & 1,000 & 7,372 & 14\\
Test & 1,000 & 7,374 & 14\\
\hline
\end{tabular}
\end{table}

In our research, we leverage the validation and test dataset of MultiWOZ 2.1 to validate and evaluate our multi-label intent classification task across different LLMs. For our few-shot approach, we do not need the training data. But for BERT based baseline approach, we utilize the complete training, validation and test data. the training data for supervised training. The original dataset contains more than 10,000 dialogue sessions, each with multiple dialogue turns. The validation and test dataset consists of 1,000 dialogue sessions each (see Table~\ref{tab:data_distribution}). We perform our initial experiments using the validation data to come up with a semi-optimized prompt. Then we use that prompt across different models to perform prediction on the test data. For this study, we evaluate the model's intent recognition capacity \textbf{only for the user utterances}. The test data consists of 7,282 user turns. However, we do use the previous system utterance in the context while prompting the models. During BERT based experiment, we perform two different set of experiments, with and without the previous utterance in the context. 

\section{Experiment Setup}
\label{sec:experiment}
\subsection{Model and Hardware Selection}
As mentioned earlier, we choose three pre-trained somoLLMs with a relatively small number of parameters ($\sim$7B). As mentioned before, the models that we choose for this experiment are Llama2-7B-hf\footnote{https://huggingface.co/meta-llama/Llama-2-70b-chat-hf}, Mistral-7B-v0.1\footnote{https://huggingface.co/mistralai/Mistral-7B-v0.1}, and Yi-6B\footnote{https://huggingface.co/01-ai/Yi-6B}. 

Our hardware consists of two NVIDIA RTX A6000, each having 48GB of VRAM. NVIDIA RTX A6000 are consumer-grade visual computing GPU for desktop workstations. Although the models we chose take much less GPU memories, the 2*48GB of VRAM allowed us to perform batch inference to make the inference process faster. We used a FP16 quantized version of the original FP32 models which reduced each model size to half. Each model’s hardware requirement is given in the Table~\ref{tab:vram_usage}. 

\begin{table}[]
\centering
\caption{VRAM Usage of Different somoLLMs Models}
\label{tab:vram_usage}
\begin{tabular}{lcccc}
\hline
\textbf{Model} & \textbf{Parameters} & \textbf{Dtype} & \textbf{VRAM Usage (MiB)} & \textbf{Creator} \\
\hline
Mistral-7B-v0.1 & 7B & FP16 & 15898 & Mistral AI\\
Llama-2-7b-hf & 7B & FP16 & 14234 & Meta AI\\
Yi-6B & 6B & FP16 & 12338 & 01.AI\\

\hline
\end{tabular}
\end{table}

\subsection{Prompt Template Selection}

Prompting is a very crucial aspect of model performance on few-shot learning in large language models~\cite{llm-sam}. However, prompts are semi-optimal and require trial and error to come up with a good format. As this is a multi-label classification task, we gave our model a system prompt by describing the role and the input format. Then we provided a set of instructions, a list of intent classes with a short single-line description of each, and some examples of the input-output format. First, we randomly choose one example per intent class from the validation set. Then we ran multiple experiments on some small portion of the validation data with different instruction wording. We added some more examples based on the results of these experiments for which the model performed poorly. In the end, we provided 20 examples altogether, each in a valid JSON format. We added a single line to describe the reasoning of the output labels. We also added instructions like ``Think step-by-step'' and ``Take a deep breath'' to enhance the model’s reasoning capability as suggested in some previous research~\cite{llm-sam}. As the prompt template is relatively long, we do not provide this in the paper content.

\subsection{Model Inference}

During inference, we use parallel inference with a batch size of 4. On the on hand, this  reduced the inference time significantly. On the other hand, it requires more VRAM. However, batch size has no impact on the model's output and one can use serial inference in case of more limited VRAM. Additionally, the model will take additional VRAM depending on the size of the input prompt.

During inference, we set the \texttt{temperature} value to $0$ for deterministic outputs. This is a parameter that controls model's creativity (randomness) in language generation. Although we instruct the model to provide the output in a valid JSON format, because of the generative nature, the model outputs need to be further parsed to get the list of predicted intents. We only consider such intents recognised by the LLM that exist in our predefined intent set. If the model hallucinates an intent, or outputs no intents, the parser returns an empty list, symbolising no intents. Each of the used somoLLMs takes roughly 300 minutes to perform inference on all 7,374 user turns in the test data. On the predicted outputs we, evaluate the model’s performance using standard multi-label classification evaluation metrics. Details on the evaluation are provided in Section~\ref{sec:results}.

\section{Evaluation Results}
\label{sec:results}

In this section, we first introduce the used evaluation metrics. Afterwards, we present our results in the few-shot setting and provide also our results using supervised learning for comparison.

\begin{table}[]
\centering
\caption{Multi-label intent classification report on user utterances of MultiWOZ 2.1 test data set}
\label{tab:consolidated_classification_report}
\begin{tabular}{|l|ccc|ccc|ccc|c|}
\hline
& \multicolumn{3}{c|}{\textbf{Llama-2-7b-hf}} & \multicolumn{3}{c|}{\textbf{Mistral-7B-v0.1}} & \multicolumn{3}{c|}{\textbf{Yi-6B}} & \\  \hline
\textbf{Intent Class} & \textbf{Prec} & \textbf{Recall} & \textbf{F-Score} & \textbf{Prec} & \textbf{Recall} & \textbf{F-Score} & \textbf{Prec} & \textbf{Recall} & \textbf{F-Score} & \textbf{Support} \\ \hline
restaurant-inform & 0.57 & 0.20 & 0.30 & 0.72 & 0.44 & \textbf{0.55} & 0.28 & 0.31 & 0.29 & 1243 \\
restaurant-request & 0.11 & 0.54 & 0.19 & 0.15 & 0.84 & \textbf{0.26} & 0.04 & 0.46 & 0.07 & 243 \\
taxi-inform & 0.66 & 0.09 & 0.16 & 0.91 & 0.20 & \textbf{0.32} & 0.39 & 0.11 & 0.18 & 351 \\
taxi-request & 0.21 & 0.81 & \textbf{0.33} & 0.16 & 0.89 & 0.27 & 0.19 & 0.58 & 0.28 & 53 \\
hotel-inform & 0.68 & 0.69 & \textbf{0.68 }& 0.73 & 0.44 & 0.55 & 0.68 & 0.28 & 0.40 & 1202 \\
hotel-request & 0.12 & 0.46 & 0.18 & 0.15 & 0.78 & \textbf{0.26 }& 0.10 & 0.24 & 0.14 & 246 \\
attraction-inform & 0.31 & 0.06 & 0.10 & 0.40 & 0.24 & \textbf{0.30} & 0.23 & 0.11 & 0.14 & 569 \\
attraction-request & 0.31 & 0.43 & 0.36 & 0.39 & 0.86 & \textbf{0.54} & 0.25 & 0.39 & 0.30 & 452 \\
train-inform & 0.47 & 0.33 & 0.39 & 0.92 & 0.25 & 0.40 & 0.67 & 0.32 &\textbf{ 0.43} & 1401 \\
train-request & 0.13 & 0.68 & 0.22 & 0.18 & 0.90 & \textbf{0.30 }& 0.13 & 0.28 & 0.18 & 345 \\
police-inform & 0.10 & 0.75 & 0.18 & 0.75 & 0.75 & \textbf{0.75 }& 0.00 & 0.00 & 0.00 & 4 \\
general-bye & 0.14 & 0.95 & 0.24 & 0.21 & 1.00 & \textbf{0.35 }& 0.10 & 0.74 & 0.17 & 105 \\
general-thank & 0.75 & 0.52 & 0.62 & 0.90 & 0.83 & \textbf{0.87} & 0.75 & 0.11 & 0.18 & 1061 \\
general-greet & 0.02 & 0.29 & 0.04 & 0.04 & 0.29 &\textbf{ 0.06} & 0.01 & 0.29 & 0.03 & 7 \\
\hline
\textbf{Micro Avg} & 0.33 & 0.41 & 0.36 & 0.40 & 0.52 & \textbf{0.45} & 0.21 & 0.27 & 0.24 & 7282 \\
\textbf{Macro Avg} & 0.33 & 0.49 & 0.29 & 0.47 & 0.62 & \textbf{0.41} & 0.27 & 0.30 & 0.20 & 7282 \\
\textbf{Weighted Avg} & 0.50 & 0.41 & 0.40 & 0.67 & 0.52 & \textbf{0.50} & 0.46 & 0.27 & 0.29 & 7282 \\
\textbf{Samples Avg} & 0.34 & 0.37 & 0.35 & 0.40 & 0.47 &\textbf{ 0.42 }& 0.19 & 0.23 & 0.20 & 7282 \\ \hline
\textbf{Hamming Loss} & \multicolumn{3}{c|}{0.1006} & \multicolumn{3}{c|}{\textbf{0.0892}} & \multicolumn{3}{c|}{0.1206} & \\
\textbf{Jaccard Similarity} & \multicolumn{3}{c|}{0.3227} & \multicolumn{3}{c|}{\textbf{0.379}} & \multicolumn{3}{c|}{0.178} & \\ \hline
\end{tabular}
\end{table}

\subsection{Evaluation Metrics}
In evaluating our model's performance of multi-label intent classification, we employed a comprehensive set of evaluation metrics which is described in the following.

\textbf{\textit{Precision}} measures the proportion of correctly identified positive instances among all instances that the model classified as positive. \textbf{\textit{Recall}} assesses the proportion of actual positives that were correctly identified. \textbf{\textit{F-Score}} harmonizes precision and recall into a single metric by calculating their harmonic mean. \textbf{\textit{Micro-Average}} aggregates the contributions of all classes to compute the average metric. In a multi-label setting, micro-averaging implies accounting for the total true positives, false negatives, and false positives, which is crucial for datasets with class imbalance. \textbf{\textit{Macro-Average}} calculates the metric independently for each class and then takes the average. \textbf{\textit{Weighted Average}} takes into account the label imbalance by weighting the metric of each class by the number of true instances for each class. The latter helps in assessing the performance in datasets where the number of items is quite different among the classes, as in the MultiWOZ~2.1 dataset.

\textbf{\textit{Hamming Loss}} is the fraction of the wrong labels to the total number of labels. In multi-label classification, it measures the fraction of labels that are incorrectly predicted. A lower Hamming Loss indicates a better classifier. In the context of this study, \textbf{\textit{Jaccard Similarity Score}}~\cite{jaccard}  provides an understanding of the overlap between the predicted and actual labels.

\begin{figure}[]
\centering
\includegraphics[width=1\linewidth]{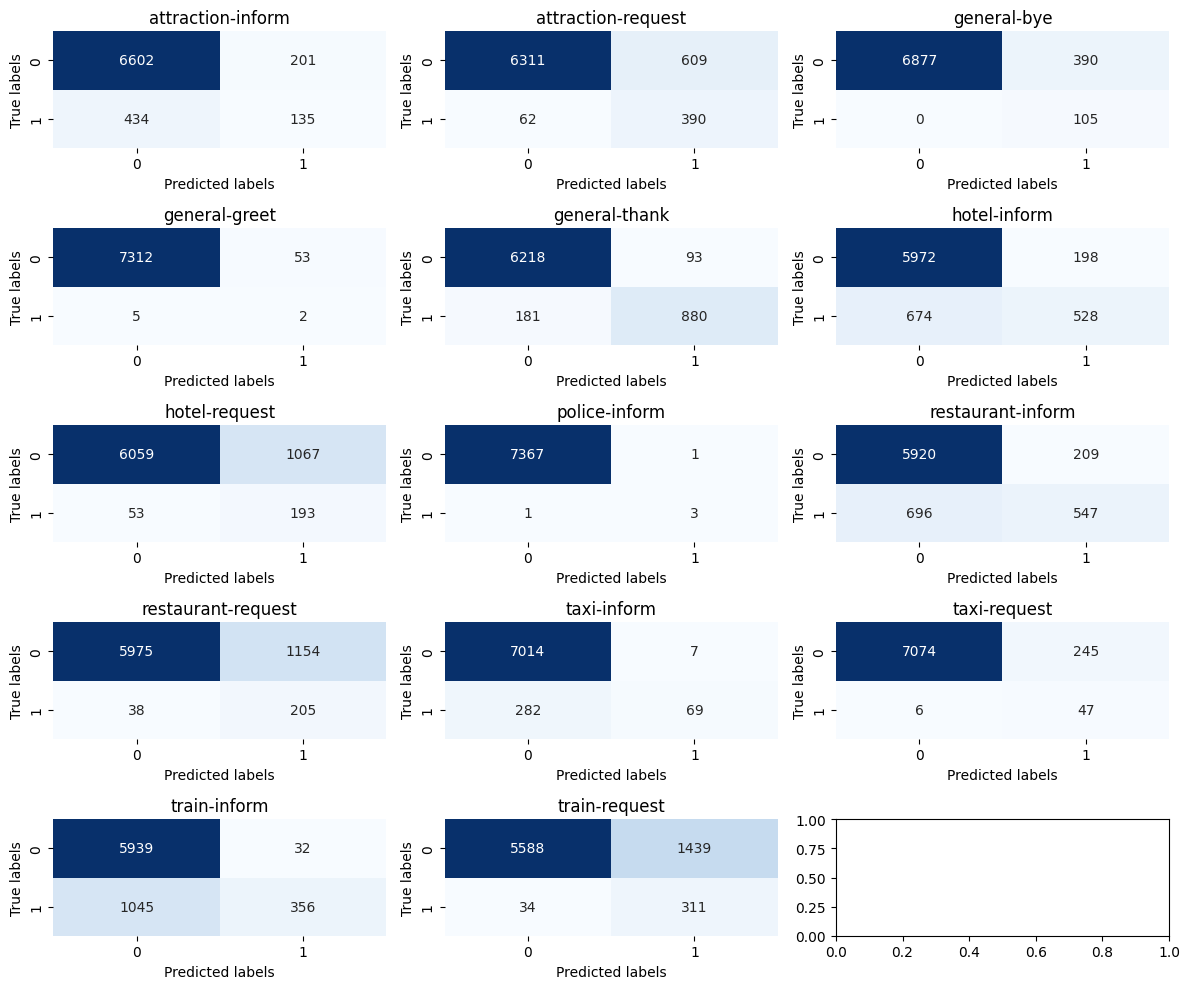}
\caption{Multi-label Intent Classification: Confusion Matrix of Mistral-7B-v0.1}
\label{fig:cm_mistral}
\end{figure}

\subsection{Evaluation results for the Few-Shot Approach}

In terms of individual classes, Mistral-7B-v0.1 outperformed the two other somoLLMs across 11 intent classes among all 14, as shown in Table~\ref{tab:consolidated_classification_report}. The model particularly performed well-detecting restaurant-inform, attraction-request, police-inform, and general-thank intents, having \textbf{$>$.5} \textbf{F-Score}. Llama-2-7b-hf performed well on the hotel-inform class and Yi-6B performed poorly across all classes, except outperforming other models detecting train-inform intents. Overall, the Mistral-7B-v0.1 outperformed two other models with a \textbf{\textit{Weighted Average} F-Score} of \textbf{0.50}. It also has a relatively lower \textbf{\textit{Humming loss}} of \textbf{0.0892} and higher \textbf{\textit{Jaccard Similarity}} of \textbf{0.397} compared to two other models. In Figure~\ref{fig:cm_mistral}, we present the Confusion Matrix for the best performing Mistral-7B-v0.1 model across different intent classes.

We can't directly compare these results with the previous works as most previous approaches report Joint Goal Accuracy (JGA) measure on the whole data set. It measures the accuracy of a system's ability to track the state (both intents and slots detection) of a dialogue across all turns. Current state-of-the-art is a BERT based architecture that has a JGA accuracy of 60.61\%~\cite{jagmetric}. For better comparison in this paper, we have evaluated a supervised learning approach using a BERT based architecture using the same dataset and task, and the results are provide in Section~\ref{subsec:eval-supervised}.

However, we intend to compare open-source LLMs' performance on multi-label intent detection with few-shot learning and to provide a framework for such comparisons. It's interesting that even if Yi-6B outperforms two other models in OpenLLMLeaderboard, it performs the least for this particular task.

One aspect of the Large Language Models is to note that they are trained on very large corpora, often containing a large amount of textual data that is available in the internet. So there is a chance that this dataset might already be part of the model's initial training data. But as the models learn to generalize rather than memorize, and we significantly changes the formatting of the original dataset, we can consider that this has few effects on the experiment. The relatively poor performance by the other two somoLLMs on this specific task 
 also supports this claim. Moreover, there is no proposed method so far to make the model forget a particular part of its training data in a few shot setup.

\subsection{Evaluation Results for Supervised Learning}
\label{subsec:eval-supervised}

\begin{table}[htbp]
\centering
\caption{Multi-label intent classification report on user utterances of MultiWOZ 2.1 test data set}
\label{tab:consolidated_classification_report}
\begin{tabular}{|l|ccc|ccc|ccc|c|}
\hline
& \multicolumn{3}{c|}{\textbf{Mistral-7B-v0.1}} & \multicolumn{3}{c|}{\textbf{BertFSC}} & \multicolumn{3}{c|}{\textbf{BertFSC+Prev.Msg.}} & \\ \hline
\textbf{Intent Class} & \textbf{Prec} & \textbf{Recall} & \textbf{F-Score} & \textbf{Prec} & \textbf{Recall} & \textbf{F-Score} & \textbf{Prec} & \textbf{Recall} & \textbf{F-Score} & \textbf{Support} \\ \hline
restaurant-inform & 0.72 & 0.44 & 0.55 & 0.89 & 0.89 & 0.89 & 0.92 & 0.92 & \textbf{0.92} & 1243 \\
restaurant-request & 0.15 & 0.84 & 0.26 & 0.61 & 0.56 & 0.58 & 0.82 & 0.89 & \textbf{0.86} & 243 \\
taxi-inform & 0.91 & 0.20 & 0.32 & 0.93 & 0.89 & \textbf{0.91} & 0.89 & 0.93 & \textbf{0.91} & 351 \\
taxi-request & 0.16 & 0.89 & 0.27 & 0.92 & 0.92 & 0.92 & 1.00 & 0.94 & \textbf{0.97} & 53 \\
hotel-inform & 0.73 & 0.44 & 0.55 & 0.92 & 0.89 & 0.90 & 0.91 & 0.91 & \textbf{0.91} & 1202 \\
hotel-request & 0.15 & 0.78 & 0.26 & 0.76 & 0.54 & 0.63 & 0.81 & 0.83 & \textbf{0.82} & 246 \\
attraction-inform & 0.40 & 0.24 & 0.30 & 0.89 & 0.90 & \textbf{0.89} & 0.89 & 0.89 & \textbf{0.89} & 569 \\
attraction-request & 0.39 & 0.86 & 0.54 & 0.84 & 0.69 & 0.76 & 0.93 & 0.91 & \textbf{0.92} & 452 \\
train-inform & 0.92 & 0.25 & 0.40 & 0.94 & 0.95 & 0.95 & 0.96 & 0.96 & \textbf{0.96} & 1401 \\
train-request & 0.18 & 0.90 & 0.30 & 0.86 & 0.79 & 0.82 & 0.84 & 0.85 & \textbf{0.84} & 345 \\
police-inform & 0.75 & 0.75 & 0.75 & 1.00 & 1.00 & \textbf{1.00} & 1.00 & 0.75 & 0.86 & 4 \\
general-bye & 0.21 & 1.00 & 0.35 & 1.00 & 1.00 & \textbf{1.00} & 1.00 & 0.99 & \textbf{1.00} & 105 \\
general-thank & 0.90 & 0.83 & 0.87 & 0.99 & 0.98 & \textbf{0.99} & 0.99 & 0.98 & 0.98 & 1061 \\
general-greet & 0.04 & 0.29 & 0.06 & 1.00 & 0.71 & \textbf{0.83} & 1.00 & 0.57 & 0.73 & 7 \\
\hline
\textbf{Micro Avg} & 0.40 & 0.52 & 0.45 & 0.91 & 0.88 & 0.89 & 0.92 & 0.93 & \textbf{0.92} & 7282 \\
\textbf{Macro Avg} & 0.47 & 0.62 & 0.41 & 0.74 & 0.69 & 0.71 & 0.76 & 0.72 & \textbf{0.74} & 7282 \\
\textbf{Weighted Avg} & 0.67 & 0.52 & 0.50 & 0.91 & 0.88 & 0.89 & 0.92 & 0.93 & \textbf{0.92} & 7282 \\
\textbf{Samples Avg} & 0.40 & 0.47 & 0.42 & 0.81 & 0.81 & 0.80 & 0.85 & 0.85 & \textbf{0.85} & 7282 \\ \hline
\end{tabular}
\end{table}

In a second experiment, we use an approach that has frequently been used for intent detection in the last years. The BERT transformer model \cite{devlin-etal-2019-bert} can be combined with a simple classification head and trained on the task of intent classification. We use the BertForSequenceClassification available on Hugging Face\footnote{https://huggingface.co} with a weighted BCEWithLogitsLoss\footnote{https://pytorch.org/docs/stable/generated/torch.nn.BCEWithLogitsLoss.html} to accommodate the multi-class problem. As a tokenizer, we used the pretrained bert-base-uncased tokenizer (110M parameters). The model is trained on the 56,778 training samples of the MultiWOZ dataset. In two approaches, we first used the user utterance on its own (BertFSC) and then ran another experiment with the last system utterances concatenated in front of the user utterance to provide more context (BertFSC+PrevMsg). We implemented early stopping with a patience of 3 epochs and set the number of epochs to 20 and the batch size to 8. The model with the best performance on the validation set was used for evaluation. This was reached at epoch 5-6 but the training itself was running the additional 3 epochs. Each epoch took 10-15 minutes to train where the single, shorter utterance was faster than the approach with both the previous system utterances and the user utterance combined (around 5 minutes faster). After training was complete, we first tested the performance on the validation set. Each intent with a score above a threshold was considered as recognised and all other intents were not considered for that sample. The threshold was picked by evaluating 9 possible thresholds and their micro f1-score on the validation set. In both experiments, the best performing threshold was 0.5 which was then applied to the final evaluation on the test set. The final evaluation time to get the results on all 7,282 test samples took less than a minute.

The results this experiment show a satisfying performance with an overall f1-score of 0.89. Additionally, adding the previous system utterance to the input improved the performance further with a final f1-score of 0.92. The full script, all results, the training, validation losses over all epochs, and the data can be found in our git repository\footnote{https://git.tu-berlin.de/goerzig/nlu/}.

\section{Discussion and Future Works}
\label{sec:discussion}

\begin{table}[]
\centering
\caption{Performance comparison of Mistral and BERT}
\label{tab:mistral_vs_bert_comparison}
\begin{tabular}{|l|c|c|}
\hline
\textbf{Category} & \textbf{Mistral} & \textbf{BERT} \\ \hline
\#TrainingSamples & \textbf{30} & 56778 \\
Training Time (min) & \textbf{0} & 80-120 \\
Evaluation Time (min) & 300 & \textbf{\textless 1} \\
\#Parameters & 7B & \textbf{110M} \\ \hline
Performance (micro f1) & 0.45 & \textbf{0.92} \\ \hline
\end{tabular}
\end{table}

In this paper, we experimented with three relatively small open-source pre-trained Large Language Models (somoLLMs) to perform few-shot multi-label classification on the MultiWOZ~2.1 dataset containing 14 intent classes and compare their performance using standard evaluation metrics. We found that Mistral-7B-v0.1 outperforms Llama2-7b-hf and Yi-6B and achieves a 50\% weighted average F-Score. However, we also found that a BERT-based approach using supervised learning still outperforms somoLLMs -- which is only applicable if sufficient training data is available. Table~\ref{tab:mistral_vs_bert_comparison} gives an overview on the amount of training data and time required for the two different approaches. We also provide a framework to perform and evaluate such experiments on multi-label intent classification in consumer hardware.

We show that without requiring much training data and hardware, one can consider using this model to perform Natural Language Understanding tasks for task-oriented dialogue systems. As fine-tuning Large Language Models has often shown significant performance improvements on specific tasks~\cite{llm-sam}, we are planning to fine-tune the best-performing model and compare the results. But then again, fine-tuning comes with the trade-off of requiring more training data and hardware compared to few-shot learning. Researchers and developers of task-oriented dialogue systems can benefit from the insights of this paper to make their choice. 

\section{Acknowledgement}
\label{sec:acknow}

The presented work has been funded by the Federal Ministry of Education and Research (Germany) and the Federal State of Berlin under grant no. 16DHBKI088 for the project USOS at Technische Universität Berlin.


\end{document}